\documentclass[10pt, a4paper, conference, compsocconf]{IEEEtran}
\ifCLASSINFOpdf
   \usepackage[pdftex]{graphicx}
\else
   \usepackage[dvips]{graphicx}
\fi
\usepackage{array}
\usepackage[cmex10]{amsmath}
\usepackage{multirow}
\usepackage{subfig}
\usepackage{epsfig}
\usepackage{amssymb}
\usepackage{comment}

\renewcommand{\IEEEbibitemsep}{0pt plus 2pt}
\makeatletter
\IEEEtriggercmd{\reset@font\normalfont\footnotesize}
\makeatother
\IEEEtriggeratref{1}

\begin{document}
%
\title{Learning Surrogate Models of Document Image Quality Metrics for \\Automated Document Image Processing}

\author{\IEEEauthorblockN{Prashant Singh, Ekta Vats and Anders Hast}
\IEEEauthorblockA{Department of Information Technology\\
Uppsala University, SE-751 05 Uppsala, Sweden\\
Email: prashant.singh@it.uu.se; ekta.vats@it.uu.se; anders.hast@it.uu.se}
}


%


\maketitle

\begin{abstract}
Computation of document image quality metrics often depends upon the availability of a ground truth image corresponding to the document. This limits the applicability of quality metrics in applications such as hyperparameter optimization of image processing algorithms that operate on-the-fly on unseen documents. This work proposes the use of surrogate models to learn the behavior of a given document quality metric on existing datasets where ground truth images are available. The trained surrogate model can later be used to predict the metric value on previously unseen document images without requiring access to ground truth images. The surrogate model is empirically evaluated on the Document Image Binarization Competition (DIBCO) and the Handwritten Document Image Binarization Competition (H-DIBCO) datasets.

\end{abstract}


%
\IEEEpeerreviewmaketitle

\section{Introduction}
\label{sec:intro}

Document image quality metrics are objective measures that enable assessment and quantification of characteristics of a given document image. Such metrics are crucial for enabling automatic document processing applications, such as fully-automatic document image binarization. Specifically, document image processing algorithms involve hyperparameters that must be optimized to achieve the best possible resulting image. Hyperparameter optimization techniques such as Bayesian optimization \cite{snoek2012practical} require formulation of an \emph{objective function} to be maximized. Document image quality metrics are natural candidates as objective functions. 

In general, document image quality is calculated by comparing the image in question to the noise-free replica of the document image, known as the ground truth reference image. There exist several popular image quality metrics in literature \cite{ye2013document}. A vast majority of the methods considered Optical Character Recognition (OCR) results as document quality metrics \cite{hale2007human, kang2014deep, nayef2015metric}. Simple techniques to measure the image quality, such as Mean Squared Error (MSE) do not suffice due to the complex and degraded nature of images. There is a need for more sophisticated methods to assess image quality. Popular document quality evaluation measures \cite{gatos2009icdar,
pratikakis2016icfhr2016} include the F-Measure, the Peak-Signal-to-Noise Ratio (PSNR), the Distance Reciprocal Distortion metric (DRD) \cite{lu2004distance}, and the Negative Rate Metric (NRM). Computation of such metrics requires a corresponding distortion-free ground truth reference image for any given document image. 

In addition to ground truth images, human opinion scores have been used as ground truth in \cite{lu2004distance,obafemi2012character,kumar2012sharpness,alaei2015document} to automatically compute the document image quality metrics. A full reference document image quality assessment technique based on texture similarity index was introduced in \cite{alaei2016document} with promising results for OCR text images. There have been recent efforts to formulate image quality metrics that are not dependent on availability of ground truths. Xu et al. \cite{xu2016no} presented a no-reference image quality metric for document image quality assessment.

However, no-reference image quality metrics, such as \cite{xu2016no}, are typically designed for document images with OCR text, and focus on specific aspects of degradations that are mostly character level distortion (e.g., noise around a character, partial or overlapping characters), and are not suitable to quantify high levels of degradations in historical handwritten texts. Machine-printed documents have simple layouts and fonts, unlike handwritten documents that have complex layouts and variability in writing style. Handwritten documents suffer from degradations such as paper stains, ink bleed-through, missing or faded data, poor contrast, warping effects, etc. that hamper document readability and pose challenges for document image processing algorithms \cite{giotis2017survey}.

Such variability and severity of degradations is better captured using ground truth based document image quality metrics such as F-Measure, PSNR and DRD. Ground truth images offer a reference point, relative to which candidate images can be ranked. This immensely helps image processing algorithms in automatically evaluating the quality of processed images.

However, the reliance on the availability of ground truth images is also severely limiting. In fact, the target domain of automated document image processing consists of ground truth generation as one of the applications. Therefore, it is impractical to have access to ground truths corresponding to previously unseen document images to be processed on-the-fly. It is possible however, to have access to a \emph{training set} of document images and corresponding ground truth images.

This work explores a novel methodology wherein document quality metric scores computed using ground truth images as reference are used to train a model that learns the relationship between the difference in image quality represented by two images, and the corresponding metric score. Given two document images - an initial image and a processed image for which the quality metric is to be computed, the trained model can be used as a \emph{surrogate} that predicts the value of the metric. Training the surrogate model is a one-time investment, and requires access to input images with corresponding ground truth images. Post training, evaluation of the surrogate model is near-instant and does not require access to ground truth image for any given test image.

This paper is organized as follows. Section \ref{sec:prob} describes the concrete problem statement. Section \ref{sec:doc} discusses various document quality metrics available in literature. Section \ref{surrogate} explains the proposed surrogate modeling approach in detail. Section \ref{sec:experiments} demonstrated the efficacy of the proposed approach on the DIBCO and H-DIBCO datasets. Section \ref{sec:discussion} discusses an alternative deep learning formulation for surrogate model training. Section \ref{sec:conclusion} concludes the paper.

\section{Problem Statement}
\label{sec:prob}

The performance of document image processing tasks such as document binarization, filtering, enhancement, text or line segmentation, and high level applications such as word spotting in a document, significantly depends on the associated hyperparameter values. In general, an automated document image processing algorithm involves automatic selection of control parameters on-the-fly. This work uses document binarization as a running example throughout the text. 

Although there exist several automated document image processing methods in literature \cite{vats2017automatic, howe2013document}, a ground truth reference image is required to tune the associated hyperparameters. For example, an automatic document image binarization method is proposed in \cite{vats2017automatic}, where Bayesian optimization is used to infer the hyperparameters on-the-fly. The value of hyperparameters is chosen such that the quality metrics corresponding to the binarized image, (such as F-Measure, PSNR etc.) are maximized, or error is minimized. However, the optimization of quality metrics such as the F-Measure, PSNR, DRD and NRM is dependent upon the availability of a ground truth reference image. This limits the applicability of such methods in real world document image processing applications.

This work explores the use of surrogate models to approximate any given document image quality metric. Let $X=\{{\bf x}_i\}_{i=1}^n$ be a set of document images comprising of $n$ images. Let $G=\{{\bf g}_i\}_{i=1}^n$ be the corresponding $n$ ground truth images. Let $P=\{{\bf p}_i\}_{i=1}^n$ be the set of processed images corresponding to $X$, obtained after processing using algorithm $A$, for example, a binarization algorithm. It is possible to compute and assign various quality metrics to ${\bf p}_i$ using ${\bf g}_i$ as a reference. Let $Y=\{y_i\}_{i=1}^n$ be a vector of values computed for any such quality metric $q$ corresponding to $P$. 

Let ${\bf x}'$ be a previously unseen test document image, with ${\bf p}'$ being the processed image obtained using a given algorithm $A$. The goal is to learn a surrogate model that can predict the value $y'$ of the metric $q$ for a given pair $({\bf x}', {\bf p}')$. Such a model will enable instant on-the-fly performance feedback for the algorithm $A$ without the availability of corresponding $g$. The model $\hat{y}$ is in effect, a \emph{surrogate} of the quality metric $q$. The following section explores popular document image quality metrics.

\section{Quality Metrics}
\label{sec:doc}
The most popularly used document image quality metrics include F-Measure, PSNR, DRD and NRM. These evaluation measures compute the image quality by comparing the document image with the corresponding ground truth reference image \cite{gatos2009icdar,
pratikakis2016icfhr2016}. 


\subsection{F-Measure}
F-Measure captures accuracy, defined as the weighted harmonic mean of Precision and Recall, 
\begin{equation} 
F-Measure = \frac{ 2 \times Recall \times Precision}{Recall + Precision },
\label{eq_fm}
\end{equation}
where $Recall=\frac{TP}{TP + FN}$ and $Precision=\frac{TP}{TP + FP}$. TP, FN and FP denote True Positives, False Negatives and False Positives, respectively.


\subsection{Peak Signal-to-Noise Ratio (PSNR)}
PSNR is a popularly used metric to measure how close an image is to another image. The higher the value of PSNR, the higher the similarity between two images. PSNR is defined via the mean squared error (MSE). Given a noise-free $M$$\times$$N$ image $I$ and its noisy approximation $K$, MSE is defined as,
\begin{equation} 
MSE = \frac{\sum_{i=1}^{M}\sum_{i=1}^{N}(I(i,j) - I'(i,j))^2}{MN},
\label{eq_mse}
\end{equation}
and PSNR is defined as,
\begin{equation} 
PSNR = 10 log \Big( \frac{C^2}{MSE} \Big),
\label{eq_psnr}
\end{equation}
where $C$ is the difference between foreground and background image.

\subsection{Distance Reciprocal Distortion metric (DRD)}
DRD is used to measure the visual distortion for all the $S$ flipped pixels in binary document images \cite{lu2004distance}, and is defined as,
\begin{equation} 
DRD = \frac{\sum\limits_{k=1}^{S}DRD_k}{NUBN},
\label{eq_drd}
\end{equation}
where $DRD_k$ is the distortion of the $k$-th flipped pixel, calculated using a $5 \times 5$ normalized weight matrix $W_{Nm}$ as,
\begin{equation} 
DRD_k = {\sum\limits_{i=-2}^{2} \sum\limits_{j=-2}^{2} |GT_k(i,j) - B_k(x,y)| \times W_{Nm}(i,j)}.
\label{eq_drdk}
\end{equation}

$DRD_k$ denotes the weighted sum of the pixels in the $5 \times 5$ block of the ground truth that differ from the centered $k$-th flipped pixel at $(x,y)$ in the binarized image. NUBN is the number of non-uniform (not all black/white pixels) $8 \times 8$ blocks in the ground truth image. 

\begin{figure*}[!t]
  \centering
  \subfloat[Surrogate model training framework.]{\includegraphics[width=4.1in]{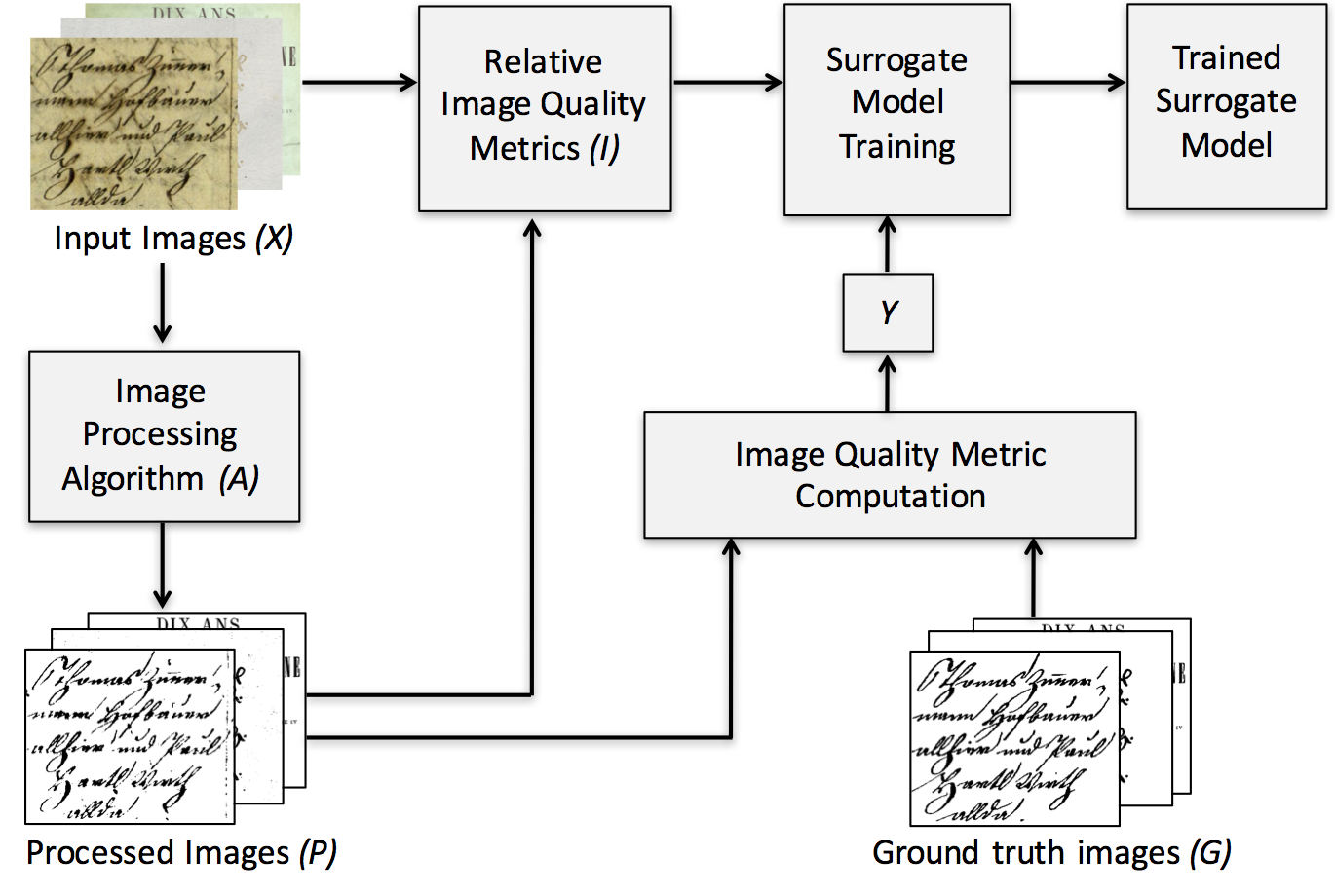}%
\label{fig1}}
\\
\subfloat[Prediction using the trained surrogate model on previously unseen data without access to ground truth images.]{\includegraphics[width=4.1in]{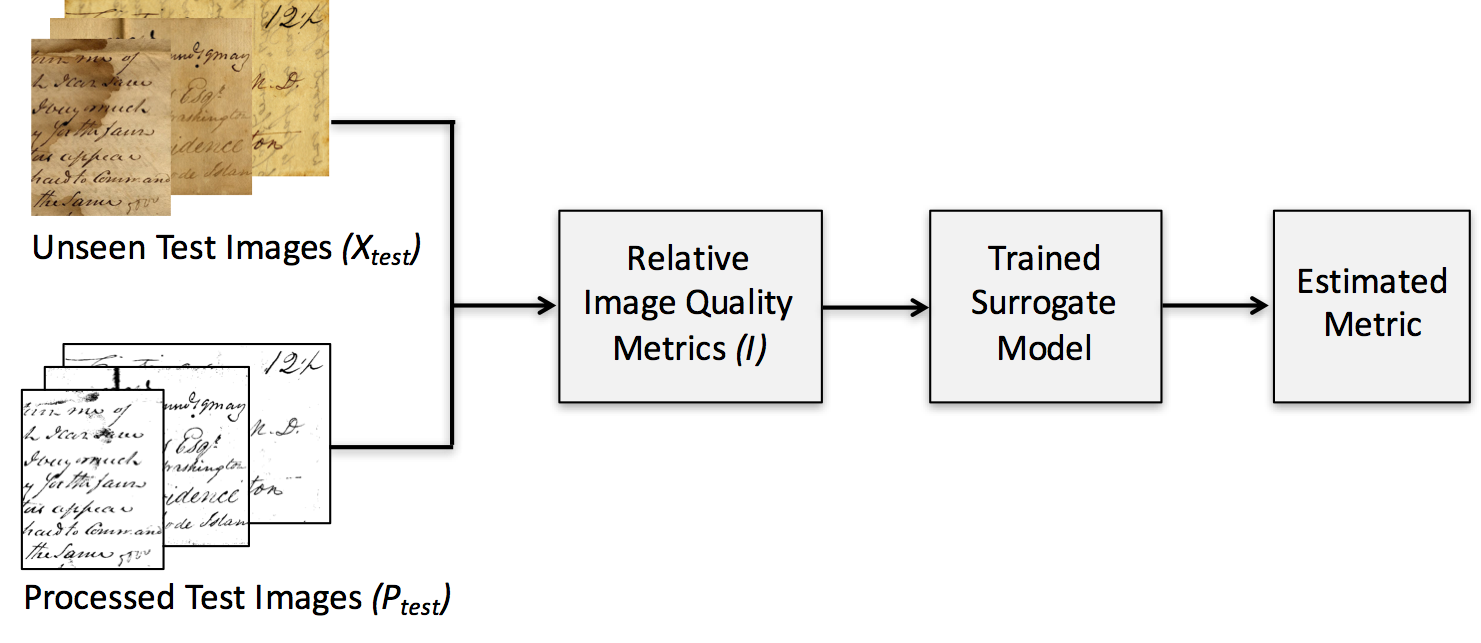}%
\label{fig2}}
\caption{Surrogate modeling framework for learning document image quality metrics.}
\label{fig_model}
\end{figure*}

\subsection{Negative Rate Metric (NRM)}
NRM measures the pixel-wise mismatch rate between the ground truth image and the resultant binarized image. NRM is defined as,
\begin{equation} 
NRM = \frac{NR_{FN} + NR_{FP}}{2},
\label{eq_nrm}
\end{equation}
where $NR_{FN}=\frac{N_{FN}}{N_{FN} + N_{TP}}$, $NR_{FP}=\frac{N_{FP}}{N_{FP} + N_{TN}}$. \newline

\noindent $NR_{FN}$ denotes the false negative rate, $NR_{FP}$ denotes the false positive rate, $N_{TP}$ is the number of true positives, $N_{FP}$ is the number of false positives, $N_{TN}$ is the number of true negatives and $N_{FN}$ is the number of false negatives. The lower the value of NRM, the better is the binarized image quality. 

%

\section{Surrogate Models for Learning Document Quality Metrics}
\label{surrogate}
Surrogate modeling \cite{gorissen2010surrogate} has emerged as a popular methodology to obtain a fast-to-evaluate approximation of a computationally expensive or data-scarce function. Since the surrogate model allows fast evaluation, it can be used in applications such as optimization, parameter space exploration and sensitivity analysis where a large number of repeated calls to the target function are required. 

For example, complex simulation codes are often used during the design process of electronic devices such as antennae, microwave filters, etc. In order to study and test the effect of varying design parameters, repeated calls to simulation codes are made. Each of these calls may take several minutes to evaluate, and this hampers the design space exploration process. Globally accurate surrogate models offer near-instant evaluation and can be used in place of such simulation codes. Obtaining such a surrogate involves preparing training data by evaluating the simulation code on a carefully selected set of parameter combinations or points, which is chosen according to a statistical design or a sampling algorithm \cite{gorissen2010surrogate}.

Automated document image processing algorithms that make use of ground truth-based image quality metrics are an excellent use-case for surrogate models. Since ground truth images are scarce, therefore it makes sense to train an accurate surrogate model of a specified image metric using the limited quantity of available ground truth images. The surrogate model can then be used to estimate the value of image quality metric on-the-fly for any input test image, and corresponding processed image.

\subsection{Surrogate Model Types}
Numerous surrogate model types exist in literature with Artificial Neural Networks (ANN), Gaussian Processes (GP) and Support Vector Machines (SVM) being popular \cite{singh2016shape}. ANNs \cite{haykin2009neural} have shown excellent results in recent years, especially in applications involving visual data, and problems involving large training sets. GPs \cite{rasmussen2006gaussian} are very popular in design optimization applications and global surrogate modeling owing to their capability of providing the variance of prediction, in addition to the prediction itself. This aids adaptive sampling algorithms in quickly searching for optima within a mathematically principled framework. 
SVR models \cite{cortes1995support} formulate the learning problem into an optimization problem that can be solved in a straightforward manner. SVR models have proven to be robust and stable in a variety of problems, and can deal with both small and large datasets. Consequently, SVR models are a reliable choice for general use in global modeling problems. This work uses ANN, GP and SVM regression models as surrogates for the purpose of experiments. However, the framework and methodology proposed herein is independent of any particular model type. A detailed discussion on the model types is out of scope in this work, and the reader is referred to \cite{smola2004tutorial,basak2007support} for SVR (support vector regression), \cite{rasmussen2006gaussian} for GPs and \cite{haykin2009neural} for ANNs.


\subsection{Model Training}
\label{approachB}
Let each document image ${\bf x}_i$ and processed image ${\bf p}_i$ be represented as a $k_1 \times k_2$ matrix. The surrogate model learns the mapping $inputs \rightarrow target$. The target is the value of the document image quality metric. The metric may also be user-defined scores. Intuitively, the inputs must represent the quantity of change or transformation the image processing algorithm $A$ has brought about in the original image ${\bf x}_i \in X$ to obtain ${\bf p}_i \in P$.
The surrogate must be able to learn the value of a given image quality metric associated with the difference and nature of transformation from ${\bf x}_i$ to ${\bf p}_i$. This transformation can also be represented as a vector of metrics that represent the differences between ${\bf x}_i$ and ${\bf p}_i$. Possible candidates to measure such transformation include the metrics explained in Section \ref{sec:doc}, e.g., F-Measure, PSNR, DRD, etc.

Let $M=\{q_j\}_{j=1}^T$ be $T$ metrics. For any given document image ${\bf x}_i$ and corresponding processed image ${\bf p}_i$, the \ $1 \times T$ vector $I_i$ represents the values of $T$ metrics as,
\begin{equation}
I_i = [q_1({\bf x}_i, {\bf p}_i),\; q_2({\bf x}_i, {\bf p}_i),\; \cdots , \; q_T({\bf x}_i, {\bf p}_i)].
\end{equation}
The complete $n \times T$  matrix $I$ represents the input variables to be learned by the surrogate model. The use of quality metrics as input variables immensely simplifies the learning problem as compared to the case of using raw images as input. The target vector $Y$ simply represents the values of a specific document image quality metric $q$ computed as,
\begin{equation}
Y_i = q({\bf p}_i, {\bf g}_i).
\end{equation}
The training set for the surrogate model is then $\mathcal{T} = (I,Y)$.
The framework of the proposed approach is pictorially described in Fig. \ref{fig_model}.

\section{Experiments}
\label{sec:experiments}

\subsection{Dataset}
The proposed surrogate-based approach is empirically evaluated on the images from seven well-known competition datasets: DIBCO 2009 \cite{gatos2009icdar}, H-DIBCO 2010 \cite{pratikakis2010h}, DIBCO 2011 \cite{gatos2011icdar2}, H-DIBCO 2012 \cite{pratikakis2012icfhr}, DIBCO 2013 \cite{pratikakis2013icdar}, H-DIBCO 2014 \cite{ntirogiannis2014icfhr2014} and H-DIBCO 2016 \cite{pratikakis2016icfhr2016}. These datasets contain machine-printed and handwritten historical document images suffering from various kinds of degradations including stained paper, faded ink or ink bleed through, wrinkles and unknown graphical symbols. In total there are 86 document images, out of which 63 randomly chosen images are used for training and 23 images for testing. As an example, the framework is applied to perform automatic image binarization using Bayesian optimization as proposed in \cite{vats2017automatic}. The document image quality metrics used as inputs for the surrogate models include PSNR, DRD and NRM. The target image quality metric to be approximated using surrogates is the F-Measure.



\subsection{Experimental Results}
The $\varepsilon$-SVR variant \cite{basak2007support} with the Sequential Minimal Optimization (SMO) \cite{platt1998sequential} solver is used for the following experiments. The hyperparameters of the SVR model are optimized using Bayesian optimization \cite{snoek2012practical}. The GP model uses a Gaussian kernel with hyperparameters being optimized using Maximum-Likelihood Estimation (MLE) \cite{rasmussen2006gaussian}. The variant of ANN used is a feed-forward back propagation neural network \cite{haykin2009neural} trained using the Levenberg-Marquardt algorithm \cite{demuth2014neural}.

The error metrics used to test the accuracy of the surrogate models are Root Relative Square Error (RRSE), Mean Absolute Error (MAE) and Root Mean Square Error (RMSE) \cite{graczyk2009comparative}.

\begin{table}[!t]
\centering
\caption{Error estimates for the surrogate models.\label{tab:resB}}
\begin{tabular}{p{1.5cm}p{1.5cm}p{1.5cm}p{1cm}}
\hline
Model Type & RRSE & MAE & RMSE \\ \hline
ANN & 0.8781 & 2.9980 & 4.1733 \\
SVR & {\bf 0.8053} & {\bf 2.7107} & {\bf 3.8272}\\
GP & 1.0633 & 3.3506 & 25.5363 \\
Ensemble & 0.8979 & 2.9477 & 5.0533 \\
\hline
\end{tabular}
\end{table}

\begin{table}[!t]
\centering
\caption{Model training and prediction times for a test dataset of $23$ document images distinct from the training set.\label{tab:times}}
\begin{tabular}{lll}
\hline
Model Type & Training Time (s) & Prediction time (s) \\ \hline
ANN & $0.3960$ & $0.0717$\\
SVR & $37.2203$ & ${\bf 0.0039}$\\
GP & ${\bf 0.1035}$ & $0.0075$\\
\hline
\end{tabular}
\end{table}

Table \ref{tab:resB} lists error estimates corresponding to the proposed model training approach described in Section \ref{approachB}. All three error measures indicate that ANN and SVR outperform GP surrogate for the given dataset. The table also contains error estimates corresponding to an ensemble model that simply averages the predictions of ANN, SVR and GP models. It can be observed from Table \ref{tab:resB} that SVR emerges as the single best performing model type.

Table \ref{tab:times} reports the time in seconds taken to train the surrogate models and the total time taken by the models to predict F-Measure values of $23$ unseen test document images. It can be seen that once the model is trained, predictions are made almost instantly. This makes the surrogate model assisted approach ideal for use on-the-fly in image processing algorithms. The time taken for preprocessing and model training is a one-time investment. A relatively high value of training time for SVR is due to the time taken to optimize hyperparameters using Bayesian optimization. This was to ensure that the hyperparameters are as close to optimal, given a relatively small training set. 


\begin{figure}[!t]
\centering
\includegraphics[width=3in]{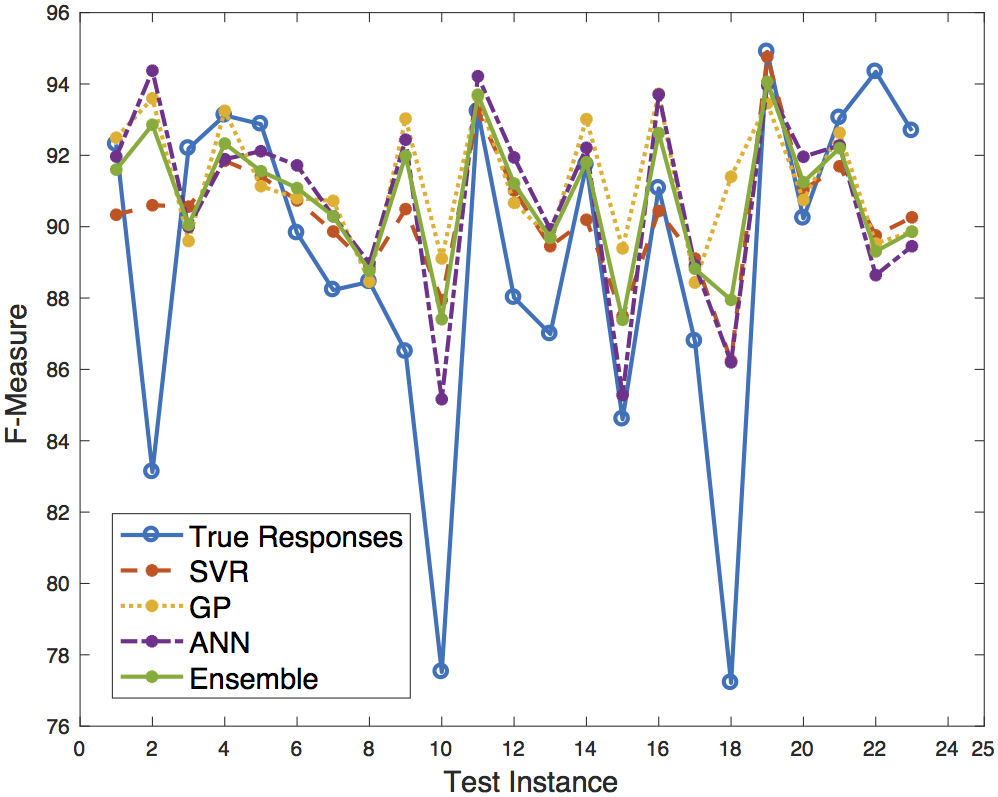}
\caption{Predicted value of F-Measure by each surrogate model type for test images. Surrogate models are accurate in general except for test instances 2, 10, 18 and 22.}
\label{fig_surrPred}
\end{figure}


Figure \ref{fig_surrPred} depicts the values of F-Measure predicted by different model types following the proposed model training approach. It can be seen that there are relatively large errors made by all model types for test instances numbers 2, 10 and 18. However, all models have been able to capture the \emph{general trend} of the test images, except for test instances 2 and 22. Even though the error is large for test instances 10 and 18, the models have been able to learn the 'downward' leaning behavior of F-Measure therein. 

\begin{figure}[!t]
\centering
\includegraphics[width=3in]{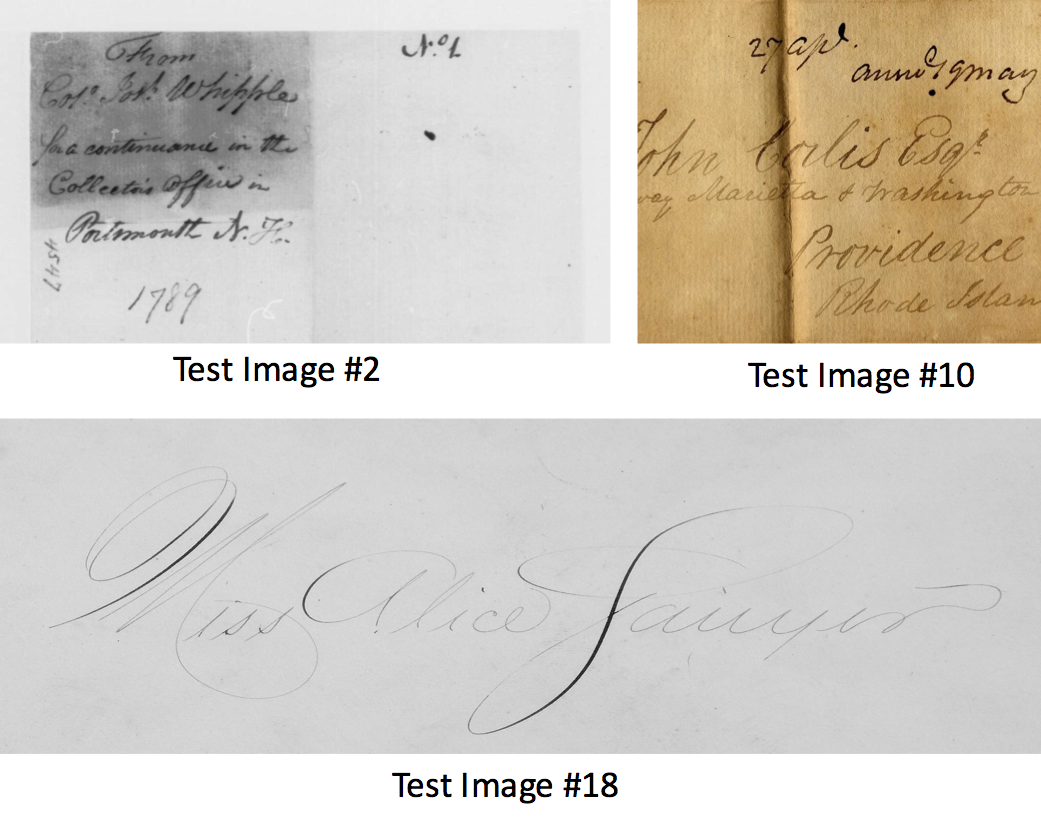}
\caption{Test images having high prediction errors.}
\label{fig_testImages}
\end{figure}

Figure \ref{fig_testImages} shows test instances 2, 10 and 18 on which all surrogate models struggled. It can be seen that test image 2 has high variation in image contrast and intensity. Test image 10 is suffering from paper wrinkles and fold marks, in addition to pen strokes of varying intensities. Test image 18 also contains variation in pen stroke intensities. Test image 22 (not shown) includes text written with multiple inks. These characteristics are not well-represented in the training set, leading to large errors in prediction of corresponding F-Measure. Having a larger training set that captures a wide variation of paper degradations, writing styles, pen stroke intensities, etc. will improve the performance of surrogate models.

\begin{figure}[!t]
  \centering
  \subfloat[Original document image.]{\includegraphics[width=3in]{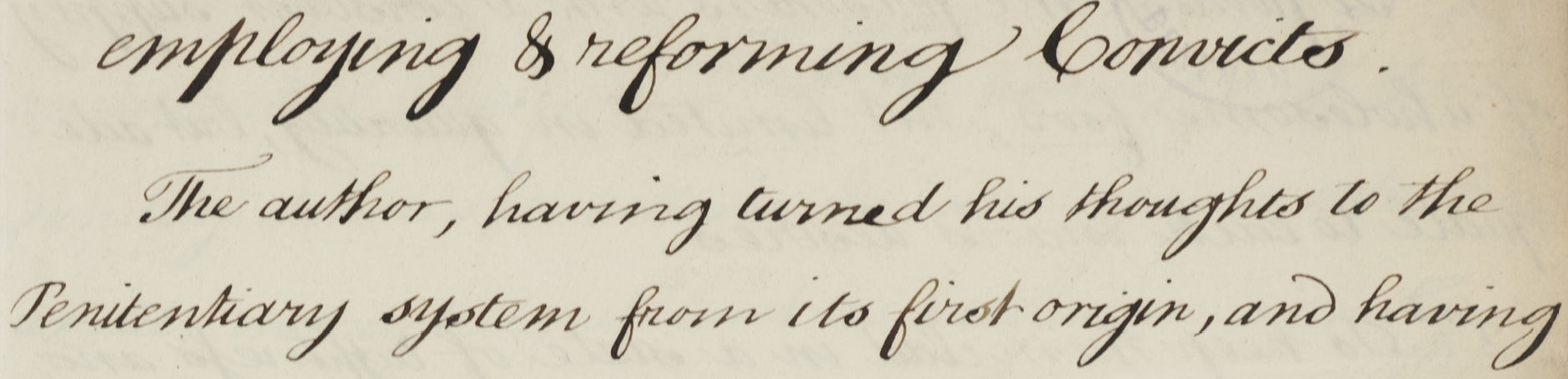}%
\label{fig_org}}
\\
\subfloat[Resultant binarized document image.]{\includegraphics[width=3in]{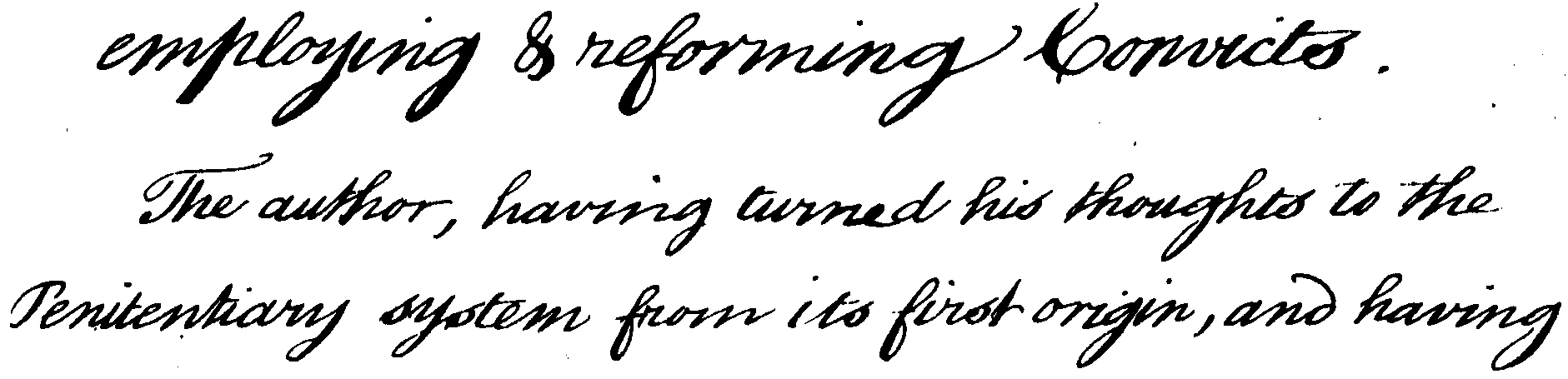}%
\label{fig_bin}}
\caption{Automatic document image binarization performed by the algorithm described in \cite{vats2017automatic}. Hyperparameters of the binarization algorithm are optimized to maximize the F-Measure approximated using the SVR surrogate model.}
\label{fig_binRes}
\end{figure}

Figure \ref{fig_binRes} shows a sample test document image binarized using the method \cite{vats2017automatic}. The hyperparameters of the binarization algorithm are optimized using Bayesian optimization \cite{snoek2012practical} as described in \cite{vats2017automatic}. The objective function to be maximized using Bayesian optimization is the F-Measure (as predicted by the SVR surrogate model trained above). The resultant image in Figure \ref{fig_bin} is clean and validates the accurate modeling of F-Measure by the SVR surrogate.

\section{Discussion: Raw Images as Input}
\label{sec:discussion}
The approach discussed herein represents the inputs as image quality metrics measuring difference between $X$ and $P$. This is done to simplify the learning problem to remain within a handful of input parameters, and allows highly efficient learning and inference.
It is also possible to consider the input and processed images themselves as input, without any post-processing to calculate quality metrics. The surrogate model will then learn the mapping $({\bf x}_i, {\bf p}_i) \rightarrow target$, where each ${\bf x}_i$ and ${\bf p}_i$ is a $k_1 \times k_2$ matrix. The representation of inputs $I: ({\bf x}_i, {\bf p}_i)$ as images is an ideal use-case of deep learning inspired surrogate models such as convolutional neural networks (CNNs) \cite{krizhevsky2012imagenet}. The caveat herein is that the training set must be sufficiently large to allow meaningful learning to proceed.

\section{Conclusion}
\label{sec:conclusion}
A novel approach is presented in this paper that uses surrogate models to learn a given document image quality metric. The surrogate model is trained on a dataset comprising of inputs that quantify differences in image quality between raw input images and corresponding processed images obtained using an image processing algorithm. The target to be approximated by the surrogate model is the value of a given document image quality metric that is computed for the training set by comparing the processed candidate images to corresponding ground truth images. Post training, the surrogate can be used to quickly predict the value of the document image quality metric for any given test pair of raw and processed document images, without any need for corresponding ground truth images. The methodology is tested on well-known publicly available document image datasets. Experimental evaluation indicates that the surrogate model is able to accurately learn the relationship between differing image quality and corresponding variation in document image quality metric value. Future work includes obtaining and experimenting with larger training sets, and exploring regression convolutional neural networks as surrogate models.


\section*{Acknowledgment}
This work was funded by the G\"oran Gustafsson foundation, the eSSENCE strategic collaboration on eScience, and the Riksbankens Jubileumsfond (Dnr NHS14-2068:1).



%


\bibliographystyle{IEEEtran}
\bibliography{refs}

\begin{thebibliography}{10}
\providecommand{\url}[1]{#1}
\csname url@samestyle\endcsname
\providecommand{\newblock}{\relax}
\providecommand{\bibinfo}[2]{#2}
\providecommand{\BIBentrySTDinterwordspacing}{\spaceskip=0pt\relax}
\providecommand{\BIBentryALTinterwordstretchfactor}{4}
\providecommand{\BIBentryALTinterwordspacing}{\spaceskip=\fontdimen2\font plus
\BIBentryALTinterwordstretchfactor\fontdimen3\font minus
  \fontdimen4\font\relax}
\providecommand{\BIBforeignlanguage}[2]{{%
\expandafter\ifx\csname l@#1\endcsname\relax
\typeout{** WARNING: IEEEtran.bst: No hyphenation pattern has been}%
\typeout{** loaded for the language `#1'. Using the pattern for}%
\typeout{** the default language instead.}%
\else
\language=\csname l@#1\endcsname
\fi
#2}}
\providecommand{\BIBdecl}{\relax}
\BIBdecl

\bibitem{snoek2012practical}
J.~Snoek, H.~Larochelle, and R.~P. Adams, ``Practical bayesian optimization of
  machine learning algorithms,'' in \emph{Advances in neural information
  processing systems}, 2012, pp. 2951--2959.

\bibitem{ye2013document}
P.~Ye and D.~Doermann, ``Document image quality assessment: A brief survey,''
  in \emph{Document Analysis and Recognition (ICDAR), 2013 12th International
  Conference on}.\hskip 1em plus 0.5em minus 0.4em\relax IEEE, 2013, pp.
  723--727.

\bibitem{hale2007human}
C.~Hale and E.~Barney-Smith, ``Human image preference and document degradation
  models,'' in \emph{Document Analysis and Recognition, 2007. ICDAR 2007. Ninth
  International Conference on}, vol.~1.\hskip 1em plus 0.5em minus 0.4em\relax
  IEEE, 2007, pp. 257--261.

\bibitem{kang2014deep}
L.~Kang, P.~Ye, Y.~Li, and D.~Doermann, ``A deep learning approach to document
  image quality assessment,'' in \emph{Image Processing (ICIP), 2014 IEEE
  International Conference on}.\hskip 1em plus 0.5em minus 0.4em\relax IEEE,
  2014, pp. 2570--2574.

\bibitem{nayef2015metric}
N.~Nayef and J.-M. Ogier, ``Metric-based no-reference quality assessment of
  heterogeneous document images,'' in \emph{SPIE 9402, Document Recognition and
  Retrieval XXII}, 2015, p. 94020L.

\bibitem{gatos2009icdar}
B.~Gatos, K.~Ntirogiannis, and I.~Pratikakis, ``Icdar 2009 document image
  binarization contest (dibco 2009),'' in \emph{Document Analysis and
  Recognition, 2009. ICDAR'09. 10th International Conference on}.\hskip 1em
  plus 0.5em minus 0.4em\relax IEEE, 2009, pp. 1375--1382.

\bibitem{pratikakis2016icfhr2016}
I.~Pratikakis, K.~Zagoris, G.~Barlas, and B.~Gatos, ``Icfhr2016 handwritten
  document image binarization contest (h-dibco 2016),'' in \emph{Frontiers in
  Handwriting Recognition (ICFHR), 2016 15th International Conference
  on}.\hskip 1em plus 0.5em minus 0.4em\relax IEEE, 2016, pp. 619--623.

\bibitem{lu2004distance}
H.~Lu, A.~C. Kot, and Y.~Q. Shi, ``Distance-reciprocal distortion measure for
  binary document images,'' \emph{IEEE Signal Processing Letters}, vol.~11,
  no.~2, pp. 228--231, 2004.

\bibitem{obafemi2012character}
T.~Obafemi-Ajayi and G.~Agam, ``Character-based automated human perception
  quality assessment in document images,'' \emph{IEEE Transactions on Systems,
  Man, and Cybernetics-Part A: Systems and Humans}, vol.~42, no.~3, pp.
  584--595, 2012.

\bibitem{kumar2012sharpness}
J.~Kumar, F.~Chen, and D.~Doermann, ``Sharpness estimation for document and
  scene images,'' in \emph{Pattern Recognition (ICPR), 2012 21st International
  Conference on}.\hskip 1em plus 0.5em minus 0.4em\relax IEEE, 2012, pp.
  3292--3295.

\bibitem{alaei2015document}
A.~Alaei, D.~Conte, and R.~Raveaux, ``Document image quality assessment based
  on improved gradient magnitude similarity deviation,'' in \emph{Document
  Analysis and Recognition (ICDAR), 2015 13th International Conference
  on}.\hskip 1em plus 0.5em minus 0.4em\relax IEEE, 2015, pp. 176--180.

\bibitem{alaei2016document}
A.~Alaei, D.~Conte, M.~Blumenstein, and R.~Raveaux, ``Document image quality
  assessment based on texture similarity index,'' in \emph{Document Analysis
  Systems (DAS), 2016 12th IAPR Workshop on}.\hskip 1em plus 0.5em minus
  0.4em\relax IEEE, 2016, pp. 132--137.

\bibitem{xu2016no}
J.~Xu, P.~Ye, Q.~Li, Y.~Liu, and D.~Doermann, ``No-reference document image
  quality assessment based on high order image statistics,'' in \emph{Image
  Processing (ICIP), 2016 IEEE International Conference on}.\hskip 1em plus
  0.5em minus 0.4em\relax IEEE, 2016, pp. 3289--3293.

\bibitem{giotis2017survey}
A.~P. Giotis, G.~Sfikas, B.~Gatos, and C.~Nikou, ``A survey of document image
  word spotting techniques,'' \emph{Pattern Recognition}, vol.~68, pp.
  310--332, 2017.

\bibitem{vats2017automatic}
E.~Vats, A.~Hast, and P.~Singh, ``Automatic document image binarization using
  bayesian optimization,'' in \emph{Proceedings of the 2017 Workshop on
  Historical Document Imaging and Processing (In Press)}.\hskip 1em plus 0.5em
  minus 0.4em\relax ACM, 2017.

\bibitem{howe2013document}
N.~R. Howe, ``Document binarization with automatic parameter tuning,''
  \emph{International Journal on Document Analysis and Recognition}, vol.~16,
  no.~3, pp. 247--258, 2013.

\bibitem{gorissen2010surrogate}
D.~Gorissen, I.~Couckuyt, P.~Demeester, T.~Dhaene, and K.~Crombecq, ``A
  surrogate modeling and adaptive sampling toolbox for computer based design,''
  \emph{Journal of Machine Learning Research}, vol.~11, no. Jul, pp.
  2051--2055, 2010.

\bibitem{singh2016shape}
P.~Singh, I.~Couckuyt, K.~Elsayed, D.~Deschrijver, and T.~Dhaene, ``Shape
  optimization of a cyclone separator using multi-objective surrogate-based
  optimization,'' \emph{Applied Mathematical Modelling}, vol.~40, no.~5, pp.
  4248--4259, 2016.

\bibitem{haykin2009neural}
S.~S. Haykin, S.~S. Haykin, S.~S. Haykin, and S.~S. Haykin, \emph{Neural
  networks and learning machines}.\hskip 1em plus 0.5em minus 0.4em\relax
  Pearson Upper Saddle River, NJ, USA:, 2009, vol.~3.

\bibitem{rasmussen2006gaussian}
C.~E. Rasmussen and C.~K. Williams, \emph{Gaussian processes for machine
  learning}.\hskip 1em plus 0.5em minus 0.4em\relax MIT press Cambridge, 2006,
  vol.~1.

\bibitem{cortes1995support}
C.~Cortes and V.~Vapnik, ``Support-vector networks,'' \emph{Machine learning},
  vol.~20, no.~3, pp. 273--297, 1995.

\bibitem{smola2004tutorial}
A.~J. Smola and B.~Sch{\"o}lkopf, ``A tutorial on support vector regression,''
  \emph{Statistics and computing}, vol.~14, no.~3, pp. 199--222, 2004.

\bibitem{basak2007support}
D.~Basak, S.~Pal, and D.~C. Patranabis, ``Support vector regression,''
  \emph{Neural Information Processing-Letters and Reviews}, vol.~11, no.~10,
  pp. 203--224, 2007.

\bibitem{pratikakis2010h}
I.~Pratikakis, B.~Gatos, and K.~Ntirogiannis, ``H-dibco 2010-handwritten
  document image binarization competition,'' in \emph{Frontiers in Handwriting
  Recognition (ICFHR), 2010 International Conference on}.\hskip 1em plus 0.5em
  minus 0.4em\relax IEEE, 2010, pp. 727--732.

\bibitem{gatos2011icdar2}
I.~Pratikakis, B.~Gatos, and K.~Ntirogiannis, ``Icdar 2011 document image
  binarization contest (dibco 2011),'' in \emph{Document Analysis and
  Recognition, 2011. ICDAR'11. 11th International Conference on}.\hskip 1em
  plus 0.5em minus 0.4em\relax IEEE, 2011, pp. 1506--1510.

\bibitem{pratikakis2012icfhr}
I.~Pratikakis, B.~Gatos, and K.~Ntirogiannis, ``Icfhr 2012 competition on
  handwritten document image binarization (h-dibco 2012),'' in \emph{Frontiers
  in Handwriting Recognition (ICFHR), 2012 International Conference on}.\hskip
  1em plus 0.5em minus 0.4em\relax IEEE, 2012, pp. 817--822.

\bibitem{pratikakis2013icdar}
I.~Pratikakis, B.~Gatos, and K.~Ntirogiannis, ``Icdar 2013 document image
  binarization contest (dibco 2013),'' in \emph{Document Analysis and
  Recognition (ICDAR), 2013 12th International Conference on}.\hskip 1em plus
  0.5em minus 0.4em\relax IEEE, 2013, pp. 1471--1476.

\bibitem{ntirogiannis2014icfhr2014}
K.~Ntirogiannis, B.~Gatos, and I.~Pratikakis, ``Icfhr2014 competition on
  handwritten document image binarization (h-dibco 2014),'' in \emph{Frontiers
  in Handwriting Recognition (ICFHR), 2014 14th International Conference
  on}.\hskip 1em plus 0.5em minus 0.4em\relax IEEE, 2014, pp. 809--813.

\bibitem{platt1998sequential}
J.~Platt, ``Sequential minimal optimization: A fast algorithm for training
  support vector machines,'' 1998.

\bibitem{demuth2014neural}
H.~B. Demuth, M.~H. Beale, O.~De~Jess, and M.~T. Hagan, \emph{Neural network
  design}.\hskip 1em plus 0.5em minus 0.4em\relax Martin Hagan, 2014.

\bibitem{graczyk2009comparative}
M.~Graczyk, T.~Lasota, and B.~Trawi{\'n}ski, ``Comparative analysis of premises
  valuation models using keel, rapidminer, and weka,'' \emph{Computational
  Collective Intelligence. Semantic Web, Social Networks and Multiagent
  Systems}, pp. 800--812, 2009.

\bibitem{krizhevsky2012imagenet}
A.~Krizhevsky, I.~Sutskever, and G.~E. Hinton, ``Imagenet classification with
  deep convolutional neural networks,'' in \emph{Advances in neural information
  processing systems}, 2012, pp. 1097--1105.

\end{thebibliography}

\end{document}